\documentclass[letterpaper, 10 pt, conference]{ieeeconf}  

\IEEEoverridecommandlockouts                              

\usepackage{booktabs}       
\usepackage{amsfonts}       
\usepackage{nicefrac}       
\usepackage{microtype}      

\usepackage{mathtools}
\usepackage[usenames, dvipsnames]{xcolor}
\usepackage{multirow}
\usepackage{algorithm}
\usepackage{algorithmic}
\usepackage{amsfonts}
\usepackage{mathtools,amssymb}
\usepackage{fixltx2e}
\usepackage{array}
\usepackage{setspace}
\usepackage{caption} 
\usepackage{bbold}
\usepackage{bm}

\usepackage{hyperref}

\newcommand{\eqn}[1]{Eqn.~#1}
\newcommand{\fig}[1]{Fig.~#1}
\newcommand{\sect}[1]{Sec.~#1}

\let\vec\bm

\usepackage{expl3}
\ExplSyntaxOn
\newcommand\latinabbrev[1]{
  \peek_meaning:NTF . {
    #1\@}%
  { \peek_catcode:NTF a {
      #1.\@ }%
    {#1.\@}}}
\ExplSyntaxOff
\def\eg{\latinabbrev{e.g}}

\definecolor{kuanfang}{RGB}{0, 0, 255}
\definecolor{yukez}{RGB}{0, 255, 0}

\title{\LARGE \bf 
\textsc{Keto}: Learning Keypoint Representations for Tool Manipulation} 

\author{Zengyi Qin$^{1, 2}$, Kuan Fang$^{1}$, Yuke Zhu$^{1, 3}$, Li Fei-Fei$^{1}$ and Silvio Savarese$^{1}$ \\ $^{1}$Stanford University \quad \quad $^{2}$Tsinghua University \quad \quad $^{3}$NVIDIA Research}

\begin{document}

\maketitle
\thispagestyle{empty}
\pagestyle{empty}

\begin{abstract}
We aim to develop an algorithm for robots to manipulate novel objects as tools for completing different task goals. 
An efficient and informative representation would facilitate the effectiveness and generalization of such algorithms.
For this purpose, we present \textsc{Keto}, a framework of learning keypoint representations of tool-based manipulation. For each task, a set of task-specific keypoints is jointly predicted from 3D point clouds of the tool object by a deep neural network. These keypoints offer a concise and informative description of the object to determine grasps and subsequent manipulation actions. The model is learned from self-supervised robot interactions in the task environment without the need for explicit human annotations. We evaluate our framework in three manipulation tasks with tool use. Our model consistently outperforms state-of-the-art methods in terms of task success rates. Qualitative results of keypoint prediction and tool generation are shown to visualize the learned representations.

\end{abstract}

\section{Introduction}

The abilities to reason about tools, to use tools for tasks, and to create new tools have been a hallmark of natural intelligence~\cite{washburn1960tools}. Building robot intelligence capable of understanding and manipulating tools has been fascinating roboticists for decades~\cite{lovchik1999robonaut}. To manipulate an object as a tool to fulfill a goal, a robot has to acquire a rich visual understanding of the object from raw sensory data and ground this understanding in complex physical interactions with the environment. In the face of object variability and sensing uncertainty in practical scenarios, a series of prior work has developed data-driven methods with the aim to learn robust tool representations for manipulation. Amid the various learning paradigms, end-to-end training of deep neural networks has become a popular choice to learning such representations~\cite{fang2018tog, holladayIROS2019, kingma2013auto, koppula2013learning}. These methods enable latent representations of tool objects to emerge as neural network layers from end-to-end learning on large-scale datasets, circumventing the need of manually designed staged pipelines and object features. However, latent representations produced by these black-box methods often lack the compactness, interpretability, and compositionality, hindering their merits for generalizable robot control. An ideal method should enjoy the representation power of neural networks while being efficient and effective for downstream manipulation tasks.

\begin{figure}[t]
    \centering
    \includegraphics[width=\linewidth]{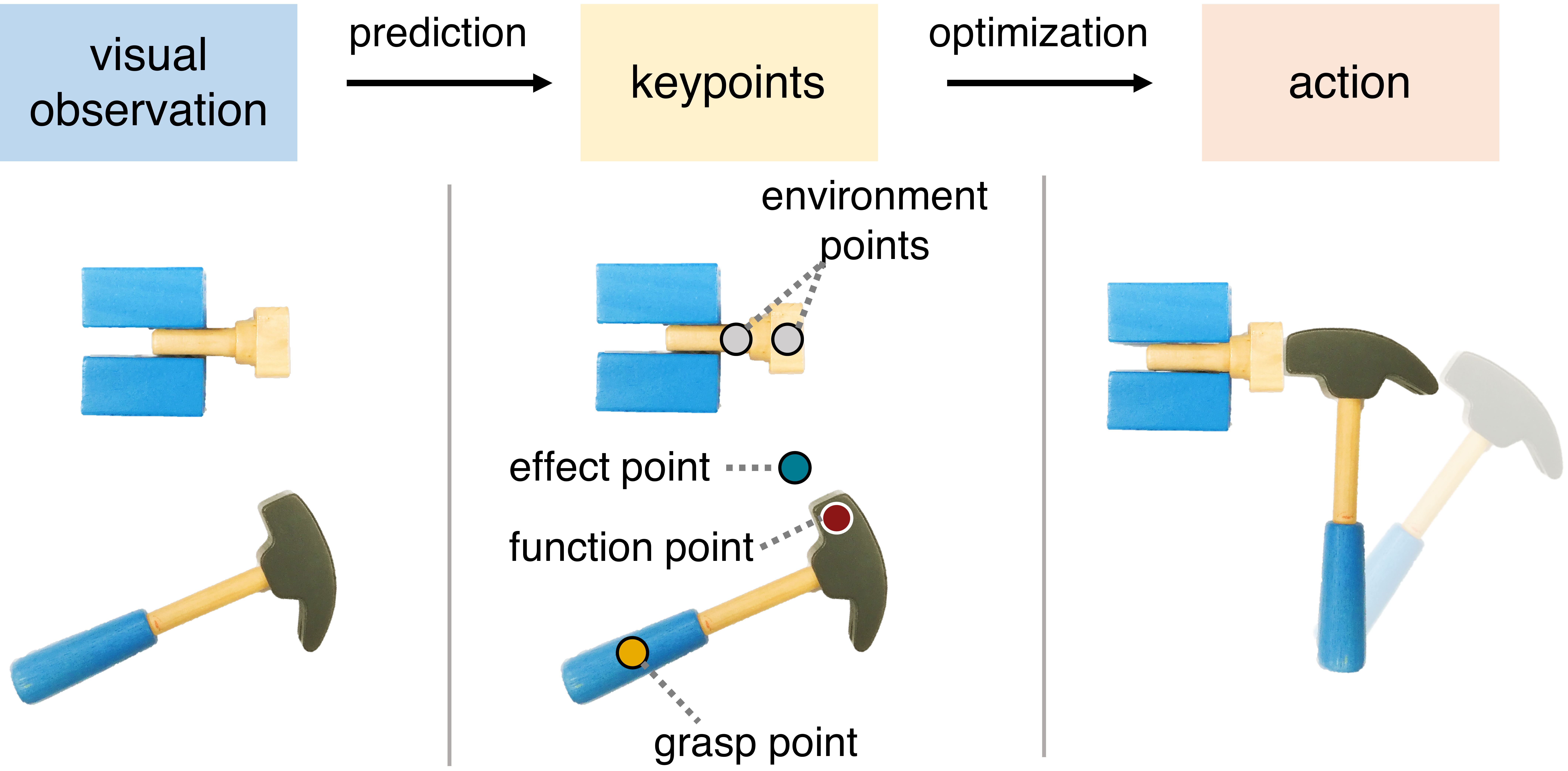}
    \caption{\textsc{Keto} uses keypoint representations for tool manipulation tasks. The tool object is represented by a set of task-specific keypoints predicted from visual observation. The keypoints of the tool and the environment together form a compact summary of the manipulation task, based on which robot motions can be effectively predicted.}
    \label{fig:keypoints}
\end{figure}

In this paper, we propose to represent the tool object as a set of keypoints to facilitate the generation of robot motion in manipulation tasks. The idea of using point-based representations has been widely explored in domains such as visual recognition~\cite{mian2008keypoint}, tracking~\cite{park2008multiple}, and reinforcement learning~\cite{jodogne2005learning}. Such representations summarize properties from high-dimensional visual data and provide structured understanding of objects. Recent works have also demonstrated the potential of using object keypoints for action optimization in manipulation tasks~\cite{manuelli2019kpam}. To achieve better generalization in contact-rich tool manipulation tasks~\cite{fang2018tog}, we would like to find keypoint representations that capture the abstract understanding of tools and suggest the optimal manipulation actions for achieving the task goal.

Pioneer work~\cite{koppula2013learning, manuelli2019kpam} on using object keypoints for robot manipulation has mostly focused on training from supervised learning on datasets with manual annotations. However, the high cost of 3D keypoint annotations severely constrains their extensibility to a broader scope. This challenge is exacerbated with the ambiguity and difficulty in defining and labeling physically grounded keypoints that best inform the manipulation tasks. Furthermore, these works predict each keypoint individually rather than in a joint optimization, which affects the quality of the predictions. Moreover, keypoints in the previous work~\cite{manuelli2019kpam} are defined at a category level, limiting the generalization capability to unseen object classes.

To this end, we propose \textsc{Keto}, a framework of learning \textbf{Ke}ypoint Representations for \textbf{To}ol Manipulation. As shown in Fig.~\ref{fig:keypoints}, we consider a task setup where a novel object is provided for the robot as a tool for solving a tool manipulation task (\eg~hammering) in the environment. In our framework, a set of keypoints is defined for each task to represent semantic information of the interactions among the robot, the tool and the environment. The tool keypoints (\emph{grasp point}, \emph{function point} and \emph{effect point}) are predicted by a task-specific keypoint generator. A set of environment points are provided beforehand to denote the task goal. Given the object keypoints and environment keypoints, the grasp and manipulation actions are produced using action optimization. In contrast to previous work~\cite{manuelli2019kpam}, the keypoint prediction is learned from self-supervised robot interactions in the task environment. No explicit annotation is needed to specify ground truth keypoints with respect to human expertise. 

We evaluate the proposed framework in three tool manipulation tasks: hammering, pushing and reaching. Each of the task requires a different strategy of grasping and manipulation for achieving the task goal. Our framework outperforms baselines using end-to-end neural network policy and hard-coded keypoints in terms of the achieved task success rate. We qualitatively demonstrate the learned keypoint representations by visualizing the predicted keypoints, as well as inversely generating the optimal tool objects given the keypoints. Video results can be found at \href{https://sites.google.com/view/ke-to}{\small\texttt{sites.google.com/view/ke-to}}


\begin{figure*}[h]
    \centering
    \includegraphics[width=\linewidth]{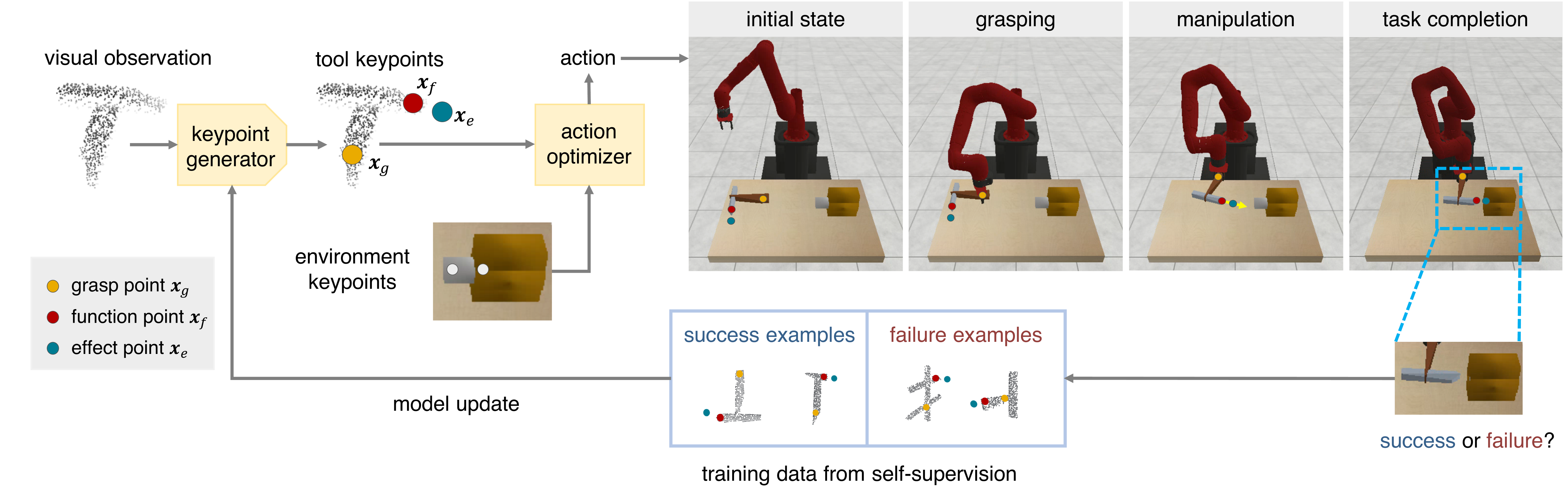}
    \caption{\textbf{Framework Overview.} The keypoint generator predicts the tool keypoints from the visual observation of the object. An action optimizer is then used to produce grasp and manipulation trajectories for the robot to achieve the task goal. Learning of the keypoint generator is conducted through trial and error in the task environment in a self-supervised manner.}
    \label{fig:pipeline}
\end{figure*}

\section{Related Work}

\subsection{Keypoint Representations}
Keypoint-based methods have been extensively studied in facial recognition~\cite{berretti20113d,mian2008keypoint,mayo20093d}, human pose estimation~\cite{belagiannis2017recurrent,carreira2016human,newell2016stacked,toshev2014deeppose, cao2017realtime, tulsiani2015viewpoints} and tracking~\cite{nebehay2014consensus, grabner2007learning, park2008multiple, hare2012efficient, chan2017robust} in computer vision and pattern recognition applications. Most keypoint representations are either defined as low-level features in detection and matching \cite{lowe2004distinctive, rublee2011orb, alahi2012freak} or used to describe object parts on a category level \cite{kar2015category, wu2016single}. 

In robotic manipulation tasks with visual inputs, keypoints can provide compact information of the environment and the objects \cite{choi2010real, finn2017deep, van2010gravity, seita2018robot, maitin2010cloth, miller2011parametrized, manuelli2019kpam}. The work most related to ours is kPAM~\cite{manuelli2019kpam} which uses category-level keypoints for manipulation tasks. We define keypoints to capture semantic information of objects and adopt a similar trajectory optimization formulation as in \cite{manuelli2019kpam}. In contrast to \cite{manuelli2019kpam}, our keypoints are task-specific and we do not assume that the object categories are known during test time. In addition, our keypoint representations are learned from self-supervised robot interactions instead of human annotations. 

\subsection{Understanding and Manipulation of Tools}
Tool use has been an essential problem for understanding intelligence in cognitive science studies \cite{beck1980animal, baber2003cognition, osiurak2010grasping, gibson2014ecological}. To enable robots to understand and manipulate tools, several pioneer works focus on predicting affordance and functional regions on tool objects \cite{lopes2007affordance, csahin2007afford, nguyen2016detecting, zhu2015understanding, kokic2017affordance, do2018affordancenet}. Manipulation of tool objects can be performed through learning and planning as shown in \cite{fang2018tog, toussaint2018differentiable, fitzgerald2019human, xie2019improvisation, holladayIROS2019}. Our work is directly related to \cite{fang2018tog} which learns to understand the synergy between grasping and manipulation for tool use from self-supervision. While most of these methods learn latent representations using end-to-end deep neural networks, our approach uses keypoint representations to provide structured and condense understanding of tool objects. Compared with \cite{fang2018tog}, our method does not rely on parameterized motion primitives and generalizes better to unseen objects during test time.


\subsection{Self-supervised Robot Learning}
Self-supervision~\cite{ridge2010self, pinto2016supersizing, levine2016learning, agrawal2016learning, finn2017deep} is used in robot learning to reduce the cost of explicit human annotations. Given the rewards received from the environment, the parameters of a policy or control system can be automatically learned through trials and errors. When collecting robot data is expensive and time-consuming in the real world, such self-supervised learning can be conducted in simulated environments \cite{bousmalis2018using, tobin2017domain, fang2018multi, fang2018tog}. Our framework follows the same practice and trains the model with self-supervised interactions in the simulated environment for each task.

\section{Problem Statement}
\label{sec:problem_statement}

We consider the problem of a robot using tools to perform manipulation tasks extended from \cite{fang2018tog}. The robot grasps a tool initially rested on the table and manipulates it to interact with the target objects in the environment. The observation of a tool is the 3D point cloud consisting of $M$ points from the robot's RGB-D camera, denoted as $\vec{o}\in\mathbb{R}^{M\times3}$. The environment and the task goal are denoted by the context $\vec{c}$. The action consists of a grasp $\vec{g}$ and the subsequent motion $\vec{a}$ of the tool. 
Let $S(\vec{o}, \vec{c}, \vec{g}, \vec{a})$ as a binary variable of task success which evaluates the outcome of taking an action $\vec{a}$ given the observed $\vec{o}$ and $\vec{c}$. Our objective is to predict the optimal grasp $\vec{g}^*$ and action $\vec{a}^{*}$ that maximizes the likelihood of task success:
\begin{equation}
\begin{aligned}
 & \vec{g}^*, \vec{a}^* = \underset{\vec{g}, \vec{a}}{\arg\max}~\Pr(S=1|\vec{o}, \vec{c}, \vec{g}, \vec{a})
\end{aligned}
\label{eqn:abstract_obj_funct}
\end{equation}

The tasks in this paper are defined by goals and constraints in terms of the spatial relationships between the tool object and the environment objects. The tool object is unseen during test time while the environment objects remain the same for each task. At the beginning of each episode, a tool object is randomly placed on the table. Under this task setup, we can use a list of spatial positions $K_{\vec{c}}$ as the environment points to summarize the environment context $\vec{c}$. 


\vspace{1mm}
\noindent \textbf{Assumptions.} We use a single-arm Sawyer robot equipped with a parallel-jaw gripper for grasping and manipulation. The depth camera is positioned top-down to acquire the point cloud observations. The observed point cloud is pre-processed by an external instance segmentation module to distinguish the tool object and the task environment context. Then $M$ points of the tool are sampled as $\vec{o}$. Following the conventions in prior work~\cite{fang2018tog}, we assume that the grasps are top-down and the tool moves on a 2D plane in parallel with the tabletop. 



\section{Method}

In this section, we describe \textsc{Keto}, a framework of learning \textbf{Ke}ypoint Representations for \textbf{To}ol Manipulation. As shown in \fig{\ref{fig:pipeline}}, our model uses a learned set of tool keypoints in each task to represent the tool object. Then the action is produced by an action optimizer given tool keypoints and environment keypoints. In the following, we will describe our keypoint representations (\sect{\ref{sect:keypoint_representation}}), the action optimizer (\sect{\ref{sect:motion_planning}}) and the keypoint generator (\sect{\ref{sect:learn_from_interact}}).

\subsection{Keypoint Representations for Tool Manipulation}
\label{sect:keypoint_representation}

Our keypoints aim to provide a compact representation of the task and guide robot motion generation for the task execution. Thus they are designed to highlight the spatial locations that well define the interactions among the robot, the tool and the environment. In our framework the keypoints in each task consists of a set of environment keypoints $K_{\vec{c}}$ and a set of tool keypoints $K_{\vec{o}}$. The environment keypoints are predefined for each task to summarize the task workspace and the tool keypoints are predicted by our model.

We assume that the tool object interacts with one or more objects as the target in a manipulation task. The environment keypoints $K_{\vec{c}}$ characterize the target and identify the force that it expects to receive. Specifically, we represent $K_{\vec{c}}=[\vec{x}_t, \vec{x}_r]$, where $\vec{x}_t$ is the target point and $\vec{x}_r$ is the receiver point. $\vec{x}_t$ denotes where the target should be in contact with the tool, or which part of the target expects a force. The vector pointing from $\vec{x}_t$ to $\vec{x}_r$ indicates the direction of that force. 

The tool keypoints $K_{\vec{o}}$ concisely summarize the tool object and guide subsequent actions. We consider tabletop tasks where the robot grasps and manipulates the tool in a continuous trajectory to achieve the task goal which is defined by the resulting motion of the target. For this purpose, three tool keypoints are considered, including a grasp point $\vec{x}_g$, a function point $\vec{x}_f$ and an effect point $\vec{x}_e$, as illustrated in \fig{\ref{fig:keypoints}}. $\vec{x}_g$ denotes the  grasping position on the object. $\vec{x}_f$ indicates the functional part of the tool object to make contact with the target. $\vec{x}_e$ is used to denote a vector pointing from $\vec{x}_f$ to $\vec{x}_e$ as the direction of force to be exerted by the tool. $\vec{x}_g$ and $\vec{x}_f$ are supposed to be predicted on the object while $\vec{x}_e$ can be chosen outside the tool.


\subsection{Grasp Selection and Motion Generation}
\label{sect:motion_planning}

The optimal grasp and manipulation actions are produced from the intermediate keypoint representations rather than raw visual observation. The grasp $\vec{g} = (\vec{x}_g, \theta_g)$ is a tuple of position $\vec{x}_g$ and orientation $\theta_g$ of the gripper in the world frame. The manipulation motion is parameterized by the final pose of the object and the effect keypoint. Our model outputs the final pose as a tuple $\vec{a}=(\vec{x}_T, \theta_T)$, where $\vec{x}_T$ and $\theta_T$ are the position and orientation of the object at the final time step $T$. Given the predicted grasp keypoint, the optimal grasp is selected from a set of robust grasp candidates as the one spatially closest to the grasp keypoint. The grasp candidates are provided by a pre-trained neural network~\cite{mousavian2019graspnet} from the point cloud of the object. The manipulation motion is produced from an explicit optimization algorithm parameterized by the keypoints.

We design an optimization algorithm for solving the manipulation action inspired by \cite{manuelli2019kpam}. Here the goal is to find the final state of the tool object, which allows it to exert a force with a given direction to the correct part of the target. For this purpose, we formulate a Quadratic Programming problem using the keypoints. For coherence, we define $\vec{x}_T$ as the final coordinate of grasp point $\vec{x}_g$, and $\theta_T$ as the angle between the \textit{x} axis and vector $\vec{x}_f - \vec{x}_g$. We choose to solve $\vec{x}_g$ and $\vec{x}_f$ rather than directly solving $\theta_T$. Let $\vec{z} = \left[x_g, y_g, x_f, y_f\right]^T$ represent the tool pose that enables the agent to exert a force with direction $\vec{\hat{e}}=\vec{x}_r-\vec{x}_t$ to target point $\vec{x}_t$. In order to solve $\vec{z}$, two conditions need to be satisfied. First, the $\vec{x}_f$ is close to $\vec{x}_t$, which equals to minimizing $\lvert\lvert \vec{x}_f - \vec{x}_t \rvert\rvert^2$. Second, the direction of force exerted by the tool should align with the force required by the target. Since $\vec{e}=\vec{x}_e-\vec{x}_f$ denotes the force exerted by the tool and $\vec{\hat{e}}$ represents the force expected by the target, we ensure that $\vec{e}$ aligns with the direction of $\hat{\vec{e}}$. This constraint is enforced by maximizing the dot product $\vec{e}^T\hat{\vec{e}}$, which is equivalent to 
\begin{equation}
\frac{\vec{v}^T\vec{z}}{||\vec{x}_f - \vec{x}_g||^2}\text{, where } \vec{v}=
 \left[
\begin{matrix}
 ~~\alpha~cos(\gamma) + \beta~sin(\gamma) \\
 -\beta~sin(\gamma) + \beta~cos(\gamma) \\
 -\alpha~cos(\gamma) - \beta~sin(\gamma) \\
 ~~\beta~sin(\gamma) - \beta~cos(\gamma)\\
\end{matrix}
 \right].
 \label{eqn:v_z_dot}
 \end{equation}
The constant $\gamma$ is defined as the angle included in $\vec{e}$ and $\vec{x}_g - \vec{x}_f$. $(\alpha, \beta)$ represents $\hat{\vec{e}}$. In practice, the division operation in \eqn{\ref{eqn:v_z_dot}} can undermine the stability of optimization, so we replace it with $\vec{v}^T\vec{z} - ||\vec{x}_f - \vec{x}_g||^2$ instead.
The optimization objective is to minimize $\lvert\lvert \vec{x}_f - \vec{x}_t \rvert\rvert^2$ and maximize $\vec{v}^T\vec{z} - ||\vec{x}_f - \vec{x}_g||^2$. It can be expressed in a quadratic form with coefficients $Q$ and $\vec{b}$ as is in
\begin{equation}
\begin{aligned}
 & \underset{\vec{z}}{\text{minimize}}
 & & f=\vec{z}^T Q \vec{z} + \vec{b}^T \vec{z}\\
& \text{subject to} & & H\vec{z} \geqslant \vec{\epsilon}
\end{aligned}
\label{eqn:objective}
\end{equation}
where
\begin{equation}
\resizebox{.9\hsize}{!}{$
Q = \left[
\begin{matrix}
1 & 0 & -1 & 0 \\
0 & 1 & 0 & -1 \\
-1 & 0 & 2 & 0 \\
0 & -1 & 0 & 2 \\
\end{matrix}
\right],  \quad
\vec{b} = \left[
\begin{matrix}
~~~~~~-\alpha\cos(\gamma) - \beta\sin(\gamma) \\
~~~~~~~~~~\beta\sin(\gamma) - \beta\cos(\gamma) \\
-2x_t + \alpha\cos(\gamma) + \beta\sin(\gamma) \\
-2y_t - \beta\sin(\gamma) + \beta\cos(\gamma) \\
\end{matrix}
\right]$}.
\end{equation}
The inequality constraints specified by $H$ and $\vec{\epsilon}$ define a valid range of solution. This way we have formulated the action optimization as Quadratic Programming in \eqn{\ref{eqn:objective}} that can be solved using CVXPY~\cite{cvxpy} in a few milliseconds. The solution $\vec{z}$ indicates the precise final pose of the tool object. At this pose, the functional part of the tool indicated by $\vec{x}_f$ can exert a force with desired direction $\vec{\hat{e}}$ to the target. 

\begin{figure}[t]
    \centering
    \includegraphics[width=0.9\linewidth]{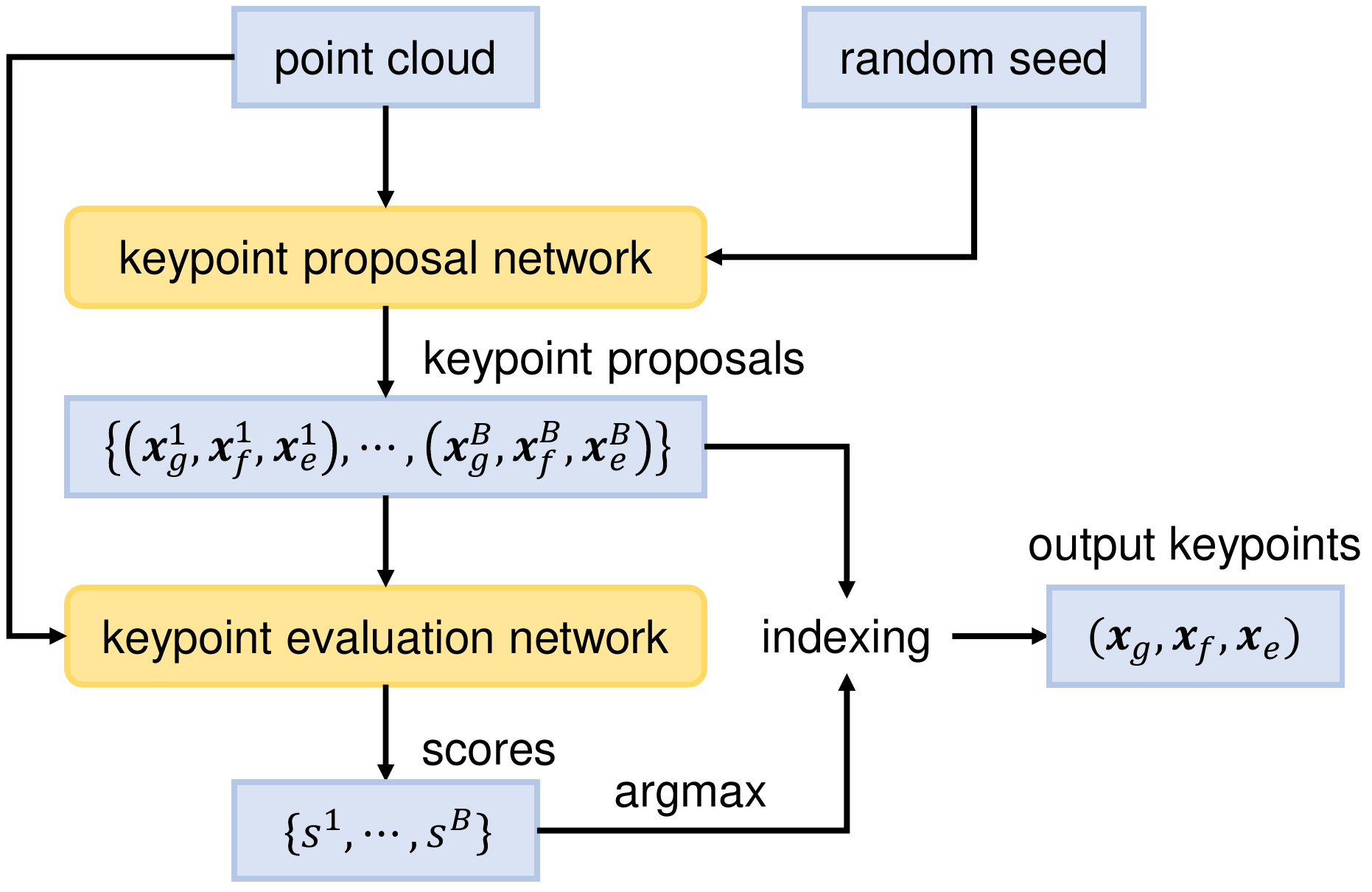}
    \caption{\textbf{Keypoint Generator.} Candidate keypoints are produced by the proposal network given the object's point cloud. The evaluation network predicts a score for each candidate and chooses the one corresponding to the highest score.}
    \label{fig:keypoint_generator}
\end{figure}

\begin{figure*}
    \centering
    \includegraphics[width=\linewidth]{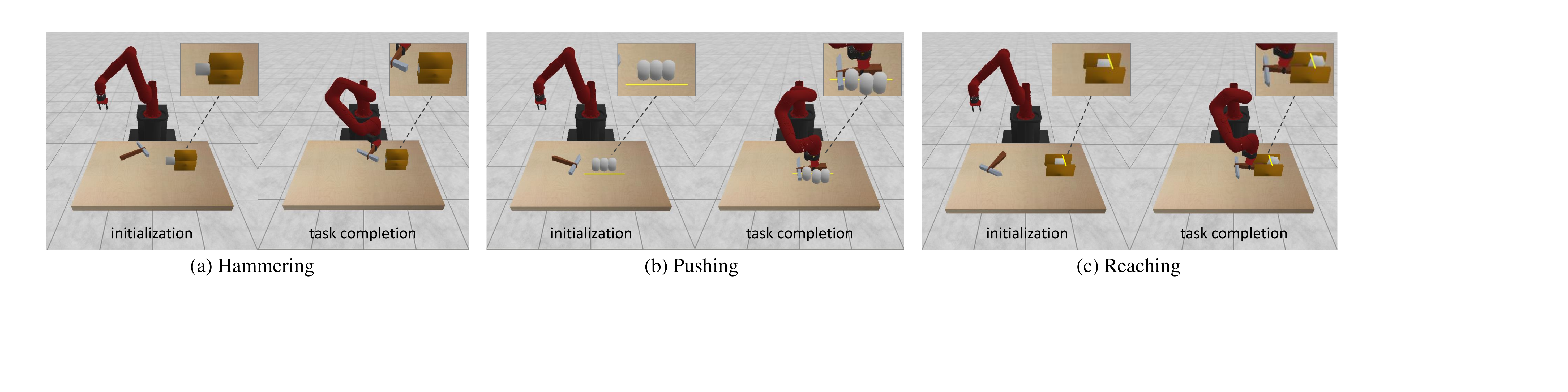}
    \caption{\textbf{Task Execution.} We show examples of tool manipulation performed using our learned model. With the same object, the model generates different grasps and motion trajectories for the robot to use the object as a tool for different purposes.}
    \label{fig:tasks}
\end{figure*}

\begin{figure}
    \centering
    \includegraphics[width=\linewidth]{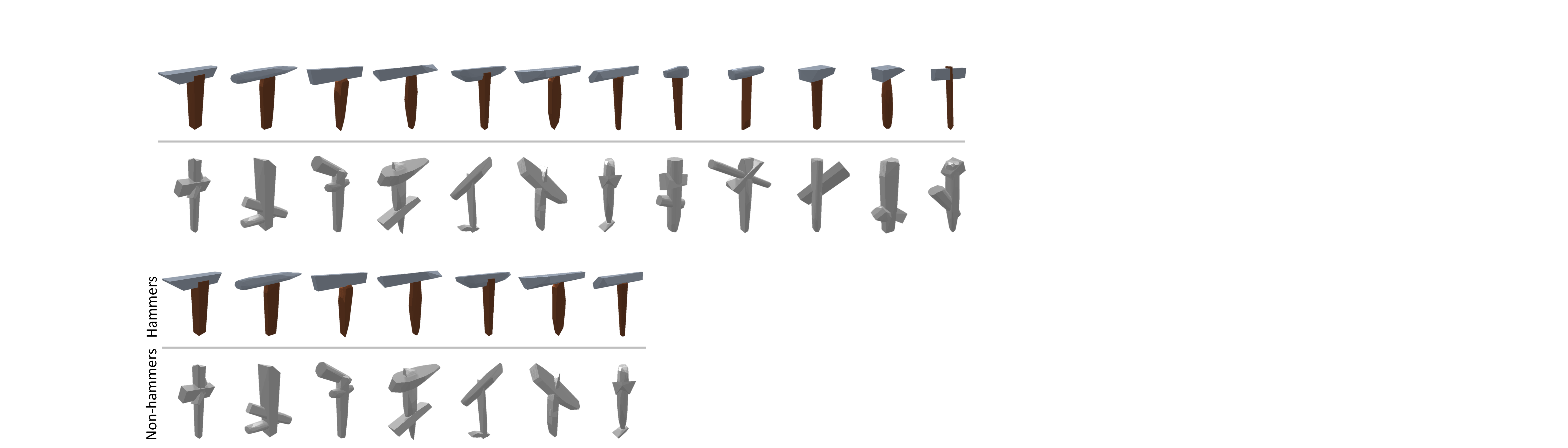}
    \caption{\textbf{Objects Examples.} Objects are categorized as hammers and non-hammers. Training and testing use different instances procedurally generated from the same distribution. 600 training objects and 600 testing objects are used. }
    \label{fig:tools}
\end{figure}
\begin{figure*}[t]
    \centering
    \includegraphics[width=\linewidth]{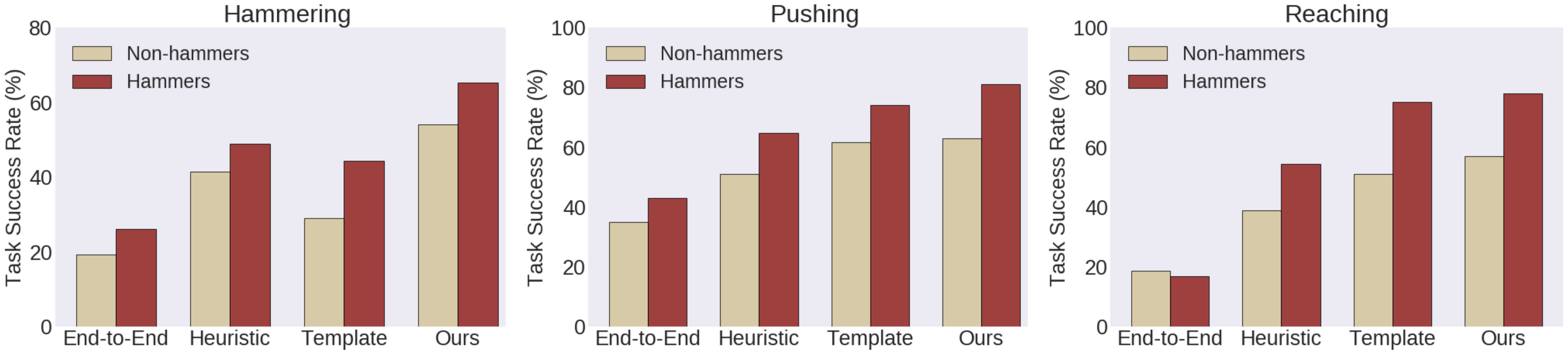}
    \caption{\textbf{Task Success Rates.} We evaluate three tool manipulation tasks using hammer-like tool objects and diverse non-hammer objects respectively. Our method consistently outperforms baselines using end-to-end neural network representations, heuristic keypoints and template matching in all scenarios. Detailed discussions can be found in \sect\ref{sec:quantitative}.}
    \label{fig:task_success_rate}
\end{figure*}

\begin{figure}[h]
    \centering
    \includegraphics[width=\linewidth]{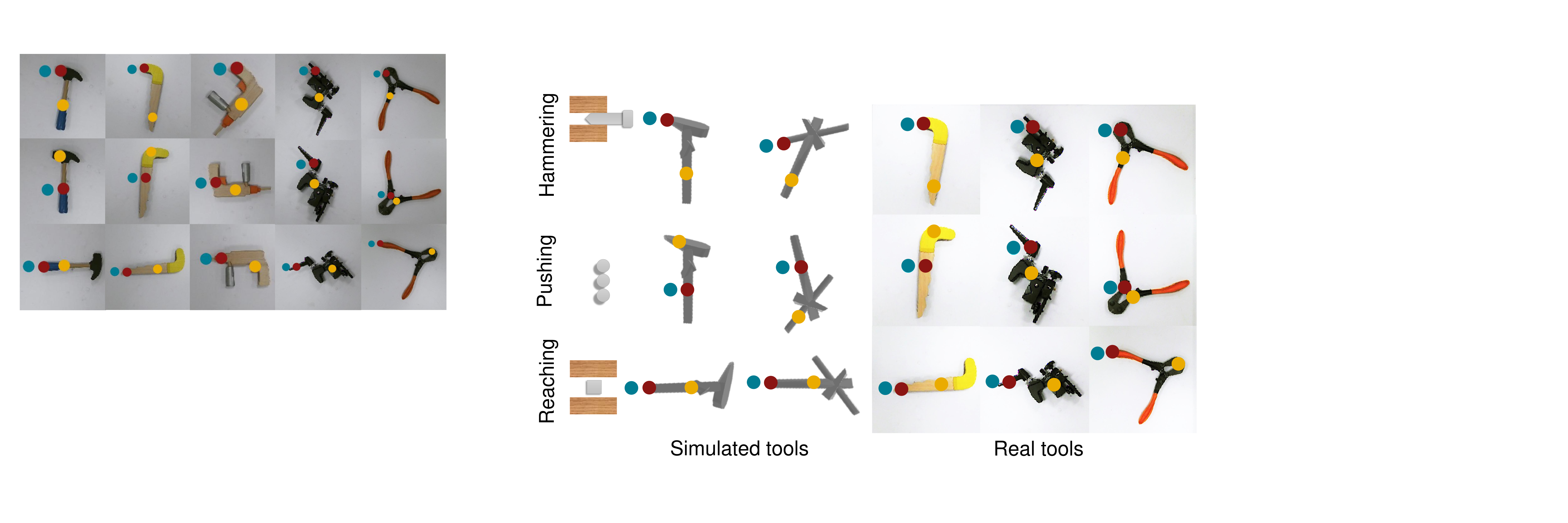}
    \caption{\textbf{Keypoint Visualization.} Predicted keypoints on simulated and real objects are overlayed with the object image. For each task, consistent patterns emerge across objects.}
    \label{fig:vis_keypoint}
\end{figure}

\begin{figure}[t]
    \centering
    \includegraphics[width=\linewidth]{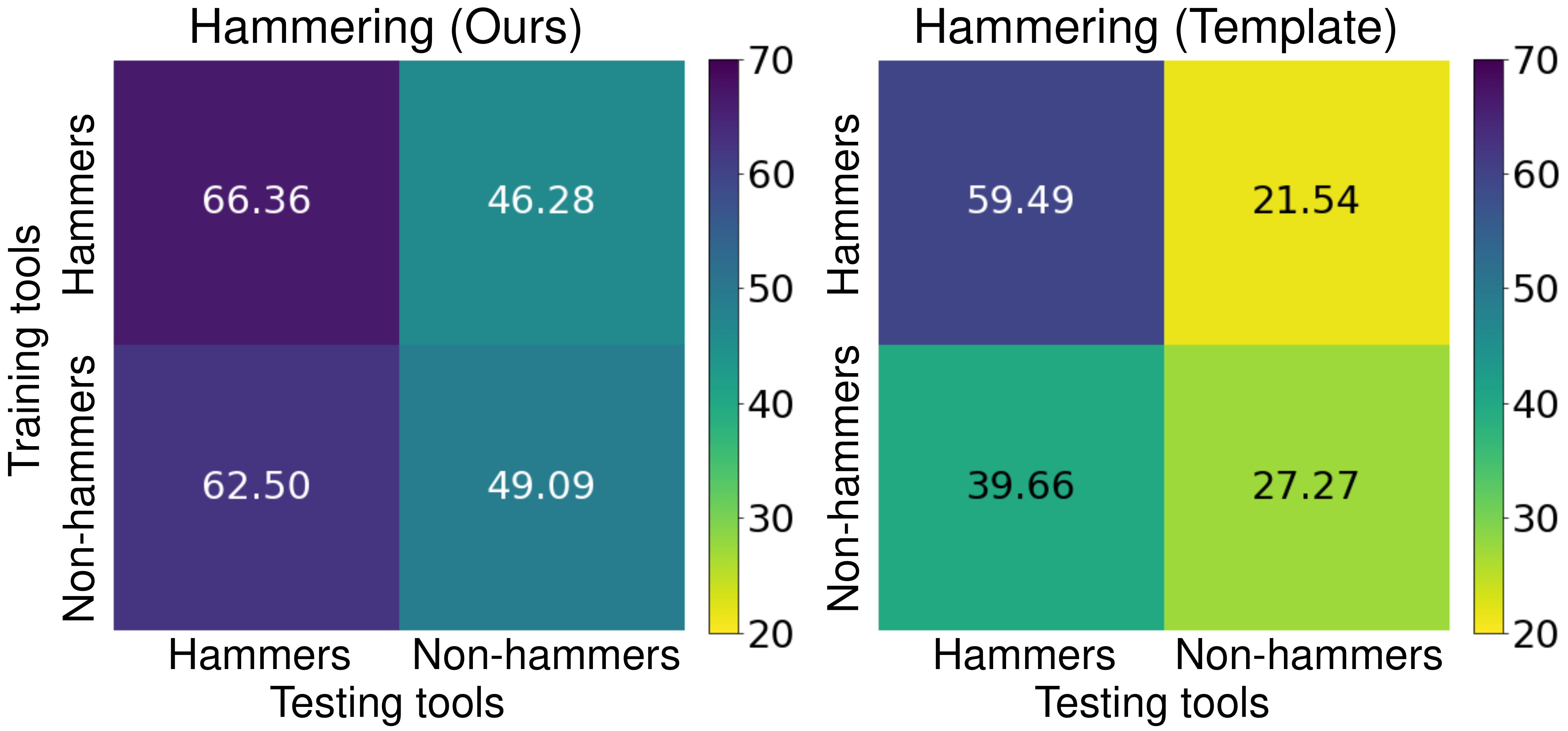}
    \caption{\textbf{Confusion Matrices across Categories.} We train and test the model on different object sets. Our method better generalizes to different categories in terms of success rates.}
    \label{fig:confuse_mat}
    \vspace{-4mm}
\end{figure}

\subsection{Keypoint Generator}
\label{sect:learn_from_interact}

Tool keypoints are predicted based on the point cloud of the object. As is shown in \fig{\ref{fig:keypoint_generator}} The keypoint generator takes the point cloud of the object as the input and outputs a tuple $K_{\vec{o}}=[\vec{x}_g, \vec{x}_f, \vec{x}_e]$. It is composed of a keypoint proposal network and a keypoint evaluation network. The keypoint proposal network produces $B = 256$ candidates using a neural network generative model \cite{mousavian2019graspnet} given the point cloud of the object and a set of random seeds. The keypoint evaluation network predicts a score for each candidate and chooses the one with the highest score as the output keypoint. Both the proposal and evaluation network use PointNet~\cite{qi2017pointnet} as the backbone feature extractor.

Learning of the keypoint generator is conducted in a self-supervised manner without explicit human annotations. As shown in \fig{\ref{fig:pipeline}}, our framework gains experience through trial and error to train the model parameters. Each trial uses a predicted set of keypoints to generate the action as described in \sect{\ref{sect:motion_planning}}. We start with generating keypoints using a heuristic policy and then switch to use the neural network model after training. After each trial, we save the point cloud, the keypoints and the task success label (if the robot robustly grasps the object and achieves the predefined task goal) as training data. The keypoint generator is encouraged to produce keypoints by following the distribution of the succeeded ones. The keypoint proposal network is trained through variational inference \cite{kingma2013auto} with a reconstruction loss which measures the $\ell_1$ distance between the succeeded keypoints and the network prediction. The keypoint evaluation network is trained with the sigmoid cross entropy loss to predict the success or failure of the subsequent manipulation given each candidate $K_{\vec{o}}$.

\section{Experiments}

The primary objective of our experiment is to examine the effectiveness of the proposed \textsc{Keto} in understanding and manipulating tools. In particular, we aim to answer four questions in the experiment: 1) How does our approach perform in manipulation tasks? 2) How can the keypoint representations improve generalization towards unseen objects? 3) Can the predicted keypoints offer an intuitive and interpretable understanding of the tool? 4) Can we use the keypoints to create novel tools?

\subsection{Experiment Setup}
\paragraph{Manipulation tasks} 
Our training and quantitative evaluation is performed in the PyBullet~\cite{coumans2016pybullet} simulation. A depth camera is mounted above the workspace to obtain a top-down view of the whole scene. We use the 3D point cloud from the camera as visual observation. We consider three tasks, including \textit{hammering}, \textit{pushing}, and \textit{reaching}, shown in Fig.~\ref{fig:tasks}. In hammering, the nail is initially half-way inside the slot. The robot grasps the tool and hammers the nail into the slot. The task succeeds if the entire nail is through the slot. In pushing, objects are placed in a row in the manipulation region. The robot uses the tool to push the objects to a given direction at the same time. After pushing, if the movement of all objects along that direction exceeds a given threshold, the task is considered completed. In reaching, one targeted object is initially in a confined tunnel. The robot has to utilize a thin part of the tool to reach and poke the target in a given direction. If its movement along that direction exceeds a threshold, the task is successfully achieved. 

\paragraph{Tool objects} The tool objects are obtained via the procedural generation approach where we construct an object by combining separate convex parts. We experiment with two object distributions consisting of hammers and non-hammers (see \fig{\ref{fig:tools}}). We generate 600 tools (300 hammers and 300 non-hammers) for training and testing respectively.

\subsection{Baseline Methods}
\paragraph{End-to-End} The End-to-End method is adopted in TOG-Net~\cite{fang2018tog} that directly predicts the manipulation actions from raw observations. The original implementation of TOG-Net utilized depth images as input. For a fair comparison, we adapt their method to taking the 3D point cloud as input with the same neural network encoder as ours.

\paragraph{Template} This is a non-parametric baseline based on template matching. Given the 3D point cloud of a test object, we search for its closest training example in terms of the Chamfer distance, and then transfer the keypoints of this training example as the keypoint representation of the test object. This baseline suffers from a significant computational burden due to template matching on a large training set.

\paragraph{Heuristic} This is a variant of our main method. Instead of learning keypoints from self-supervised interaction, we utilize a handcrafted algorithm to generate keypoint candidates. First, we apply the RANSAC algorithm to the point cloud and find the main part of the tool. We use this part to determine the grasp point $\vec{x}_g$. Second, we cluster the point cloud in groups and find the potential function point $\vec{x}_f$. Third, the effect point $\vec{x}_e$ can be estimated from $\vec{x}_f$ and the main part. As the keypoints are obtained via a handcrafted heuristic algorithm, they can lead to weaker generalization capabilities towards unseen objects.

\subsection{Data Collection and Training}
In each episode, a tool is randomly selected from the training set and placed in an arbitrary pose on the tabletop. We use our model to predict a set of keypoints $K_{\vec{o}}=[\vec{x}_g, \vec{x}_f, \vec{x}_e]$ that is fed into the action optimizer to determine the action, which is executed in simulation. If the task succeeds, $K_{\vec{o}}$ is labeled as a positive example and vice versa. To bootstrap the self-supervised learning with an initial model, we use the handcrafted algorithm from the \textit{Heuristic} baseline to increase the ratio of positive examples in the first round. We collected approximately 100K tuples of keypoints and the associated point cloud inputs for each task and trained the network for 120K iterations with batch size 128 and learning rate $10^{-4}$ using the Adam~\cite{kingma2014adam} optimizer. We trained a separate neural network model for each task respectively.

\subsection{Quantitative Results}
\label{sec:quantitative}
\paragraph{Task success rate} We present the task success rate in \fig{\ref{fig:task_success_rate}}. Our proposed model consistently outperforms the compared method in various task scenarios and evaluation criteria. The hammer tools generally lead to a higher task success rate than non-hammer tools, which we hypothesize is due to the regular and simple geometry of hammers. Compared to the end-to-end learning approach, the keypoint-based methods have attained a substantial advantage through the construct of this compact representation and the use of action optimization.


\paragraph{Generalization ability} Next we examine if our model trained on one distribution of tools (e.g., hammers) generalizes well to another distribution (e.g., non-hammers). We present the results in \fig{\ref{fig:confuse_mat}} and compare with the \textit{Template} method. It is shown that our method has a promising performance even with unseen tools. 

\subsection{Qualitative Results}
\paragraph{Keypoint prediction} Our keypoint representation is inherently interpretable. We visually examine our model's predictions on tool observations in both simulation and real environments. \fig{\ref{fig:vis_keypoint}} illustrates the predictions on a set of unseen tools for the three tasks. Our model can generate task-specific keypoints suited for different task scenarios with hammer-like objects and other irregular shapes. The real images are collected with a Kinect RGB-D camera. This result indicates that our model, trained only with synthetic data, transfers well to the real-world setup, opening up the potential of deploying this method on physical hardware.

We also demonstrate multi-stage tool use using the learned model. See qualitative results from our website: \href{https://sites.google.com/view/ke-to}{\small\texttt{https://sites.google.com/view/ke-to}}. 

\begin{figure}
    \centering
    \includegraphics[width=\linewidth]{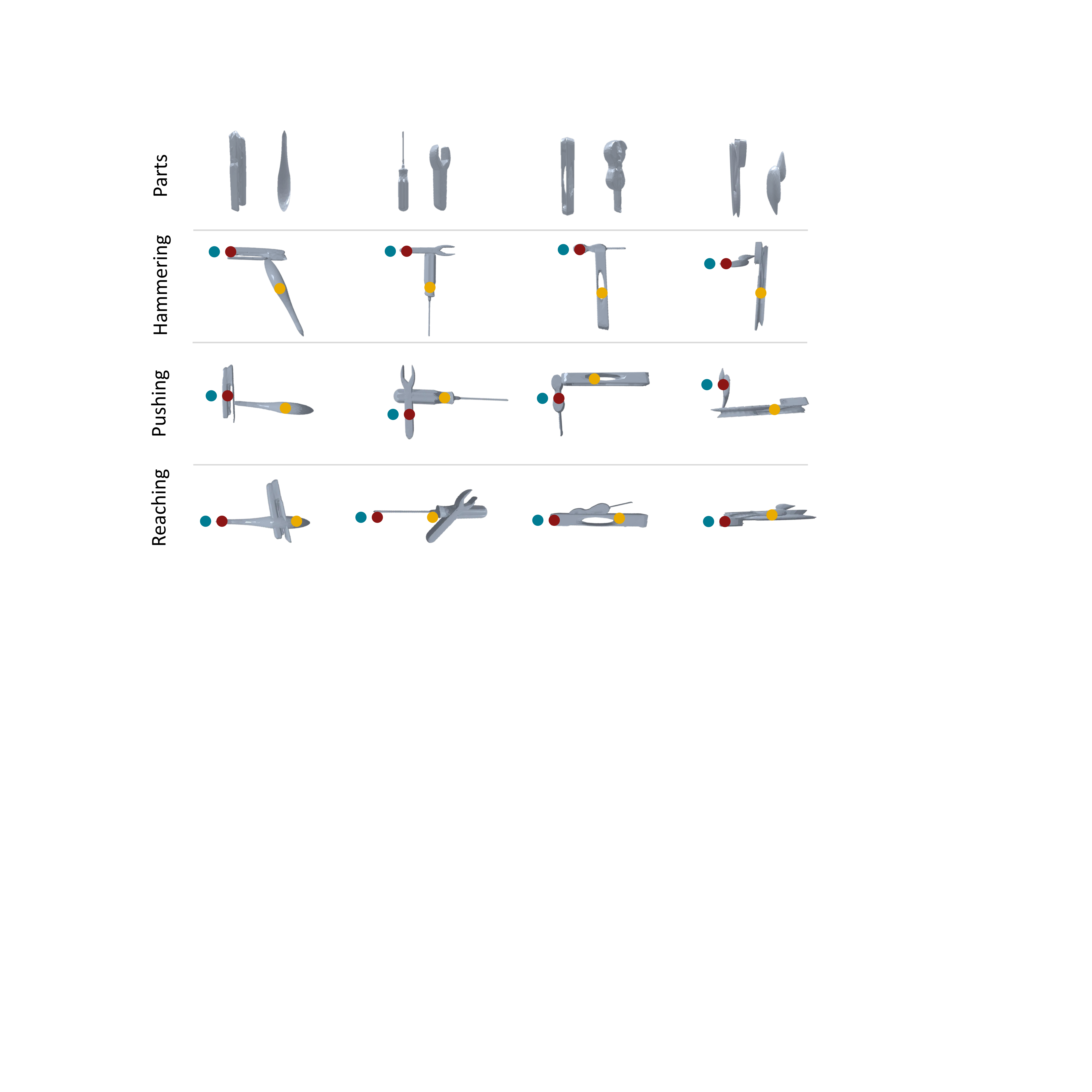}
    \caption{\textbf{Tool Generation} Given the keypoints and random object parts, optimal tool objects can be inversely composed by our model for each task by gradient ascent.}
    \label{fig:tool_gen}
\end{figure}

\paragraph{Tool generation.} The above experiments have demonstrated how to use our keypoint representations to manipulate existing tools. Moving on, we also study the inverse problem of keypoint prediction: creating new tools using the keypoints as the scaffold. Given the keypoints, we can use our model to reason about what the tool should look like and generate new tools by composing several object parts. First, we randomly choose two object parts~(see the first row of~\fig{\ref{fig:tool_gen}}) and render their point cloud respectively. The point cloud of each part is treated as a rigid body and parameterized by its translation and rotation. Second, we concatenate the point cloud of each part to obtain the whole point cloud of the object. Third, we feed the desired keypoints and the point cloud to the evaluation network in \fig{\ref{fig:keypoint_generator}} that outputs a confidence score. Forth, we increase the score by applying gradient ascent on the translation and rotation of the point cloud of each object part. When the translations and rotations of all the parts converge, the originally separated parts are combined as a tool as is shown in \fig{\ref{fig:tool_gen}}. 

\section{Conclusion}
We proposed a keypoint representation of tool manipulation, which is compact, effective, and interpretable. The keypoints are learned via self supervision generated by the robot interacting with the environment, eliminating the need of manual annotation. Our experiments have shown that using keypoints as an intermediate representation outperforms a series of competitive baselines, including end-to-end visuomotor learning. We further demonstrate the inherent interpretability of the keypoints as a compact structured summary of a tool object and illustrated its utility in creating novel tools from random object parts.  For future work, we would like to integrate keypoints with other hybrid representations to encode the semantic, geometric, and physical information of objects and environments necessary for more challenging manipulation tasks. We are also interested in extending the action optimizer to multi-step trajectory planning for long-horizon manipulation tasks instead of the single-step motion generation employed in our model. We plan to explore more sophisticated generative models for tool creation, which we believe is an exciting new research frontier. Our experiments have been primarily evaluated in physical simulation. We would like to deploy our method in physical robot hardware and evaluate its effectiveness in real-world execution.




\section*{Acknowledgement}
We acknowledge the support of the Chinese Undergraduate Visiting Research (UGVR) Program. We thank Ming Luo for her time and effort in co-organizing UGVR. We thank Roberto Mart{\'\i}n-Mart{\'\i}n for his constructive advice in problem formulation. We thank Toyota Research Institute (TRI) for providing funds to assist authors with their research. This article solely reflects the opinions and conclusions of its authors but not TRI or any other Toyota entity.

{\small
\bibliographystyle{ieee}
\bibliography{reference}

\begin{thebibliography}{10}\itemsep=-1pt

\bibitem{agrawal2016learning}
P.~Agrawal, A.~V. Nair, P.~Abbeel, J.~Malik, and S.~Levine.
\newblock Learning to poke by poking: Experiential learning of intuitive
  physics.
\newblock In {\em Advances in Neural Information Processing Systems}, pages
  5074--5082, 2016.

\bibitem{alahi2012freak}
A.~Alahi, R.~Ortiz, and P.~Vandergheynst.
\newblock Freak: Fast retina keypoint.
\newblock In {\em 2012 IEEE Conference on Computer Vision and Pattern
  Recognition}, pages 510--517. Ieee, 2012.

\bibitem{baber2003cognition}
C.~Baber.
\newblock {\em Cognition and tool use: Forms of engagement in human and animal
  use of tools}.
\newblock CRC Press, 2003.

\bibitem{beck1980animal}
B.~B. Beck.
\newblock {\em Animal tool behavior: The use and manufacture of tools by
  animals}.
\newblock Garland STPM Press New York, 1980.

\bibitem{belagiannis2017recurrent}
V.~Belagiannis and A.~Zisserman.
\newblock Recurrent human pose estimation.
\newblock In {\em 2017 12th IEEE International Conference on Automatic Face \&
  Gesture Recognition (FG 2017)}, pages 468--475. IEEE, 2017.

\bibitem{berretti20113d}
S.~Berretti, B.~B. Amor, M.~Daoudi, and A.~Del~Bimbo.
\newblock 3d facial expression recognition using sift descriptors of
  automatically detected keypoints.
\newblock {\em The Visual Computer}, 27(11):1021, 2011.

\bibitem{bousmalis2018using}
K.~Bousmalis, A.~Irpan, P.~Wohlhart, Y.~Bai, M.~Kelcey, M.~Kalakrishnan,
  L.~Downs, J.~Ibarz, P.~Pastor, K.~Konolige, et~al.
\newblock Using simulation and domain adaptation to improve efficiency of deep
  robotic grasping.
\newblock In {\em 2018 IEEE International Conference on Robotics and Automation
  (ICRA)}, pages 4243--4250. IEEE, 2018.

\bibitem{cao2017realtime}
Z.~Cao, T.~Simon, S.-E. Wei, and Y.~Sheikh.
\newblock Realtime multi-person 2d pose estimation using part affinity fields.
\newblock In {\em Proceedings of the IEEE Conference on Computer Vision and
  Pattern Recognition}, pages 7291--7299, 2017.

\bibitem{carreira2016human}
J.~Carreira, P.~Agrawal, K.~Fragkiadaki, and J.~Malik.
\newblock Human pose estimation with iterative error feedback.
\newblock In {\em Proceedings of the IEEE conference on computer vision and
  pattern recognition}, pages 4733--4742, 2016.

\bibitem{chan2017robust}
S.~Chan, X.~Zhou, and S.~Chen.
\newblock Robust adaptive fusion tracking based on complex cells and keypoints.
\newblock {\em IEEE Access}, 5:20985--21001, 2017.

\bibitem{choi2010real}
C.~Choi and H.~I. Christensen.
\newblock Real-time 3d model-based tracking using edge and keypoint features
  for robotic manipulation.
\newblock In {\em 2010 IEEE International Conference on Robotics and
  Automation}, pages 4048--4055. IEEE, 2010.

\bibitem{coumans2016pybullet}
E.~Coumans and Y.~Bai.
\newblock Pybullet, a python module for physics simulation for games, robotics
  and machine learning.
\newblock {\em GitHub repository}, 2016.

\bibitem{cvxpy}
S.~Diamond and S.~Boyd.
\newblock {CVXPY}: A {P}ython-embedded modeling language for convex
  optimization.
\newblock {\em Journal of Machine Learning Research}, 17(83):1--5, 2016.

\bibitem{do2018affordancenet}
T.-T. Do, A.~Nguyen, and I.~Reid.
\newblock Affordancenet: An end-to-end deep learning approach for object
  affordance detection.
\newblock In {\em 2018 IEEE international conference on robotics and automation
  (ICRA)}, pages 1--5. IEEE, 2018.

\bibitem{fang2018multi}
K.~Fang, Y.~Bai, S.~Hinterstoisser, S.~Savarese, and M.~Kalakrishnan.
\newblock Multi-task domain adaptation for deep learning of instance grasping
  from simulation.
\newblock In {\em 2018 IEEE International Conference on Robotics and Automation
  (ICRA)}, pages 3516--3523. IEEE, 2018.

\bibitem{fang2018tog}
K.~Fang, Y.~Zhu, A.~Garg, A.~Kuryenkov, V.~Mehta, L.~Fei-Fei, and S.~Savarese.
\newblock Learning task-oriented grasping for tool manipulation from simulated
  self-supervision.
\newblock {\em Robotics: Science and Systems (RSS)}, 2018.

\bibitem{finn2017deep}
C.~Finn and S.~Levine.
\newblock Deep visual foresight for planning robot motion.
\newblock In {\em 2017 IEEE International Conference on Robotics and Automation
  (ICRA)}, pages 2786--2793. IEEE, 2017.

\bibitem{fitzgerald2019human}
T.~Fitzgerald, E.~Short, A.~Goel, and A.~Thomaz.
\newblock Human-guided trajectory adaptation for tool transfer.
\newblock In {\em Proceedings of the 18th International Conference on
  Autonomous Agents and MultiAgent Systems}, pages 1350--1358. International
  Foundation for Autonomous Agents and Multiagent Systems, 2019.

\bibitem{gibson2014ecological}
J.~J. Gibson.
\newblock {\em The ecological approach to visual perception: classic edition}.
\newblock Psychology Press, 2014.

\bibitem{grabner2007learning}
M.~Grabner, H.~Grabner, and H.~Bischof.
\newblock Learning features for tracking.
\newblock In {\em 2007 IEEE Conference on Computer Vision and Pattern
  Recognition}, pages 1--8. IEEE, 2007.

\bibitem{hare2012efficient}
S.~Hare, A.~Saffari, and P.~H. Torr.
\newblock Efficient online structured output learning for keypoint-based object
  tracking.
\newblock In {\em 2012 IEEE Conference on Computer Vision and Pattern
  Recognition}, pages 1894--1901. IEEE, 2012.

\bibitem{holladayIROS2019}
R.~Holladay, T.~Lozano-Perez, and A.~Rodriguez.
\newblock Force-and-motion constrained planning for tool use.
\newblock In {\em International Conference on Intelligent Robots and Systems
  (IROS)}, 2019.

\bibitem{jodogne2005learning}
S.~Jodogne and J.~Piater.
\newblock Learning, then compacting visual policies.
\newblock In {\em 7th European Workshop on Reinforcement Learning}, pages
  8--10, 2005.

\bibitem{kar2015category}
A.~Kar, S.~Tulsiani, J.~Carreira, and J.~Malik.
\newblock Category-specific object reconstruction from a single image.
\newblock In {\em Proceedings of the IEEE conference on computer vision and
  pattern recognition}, pages 1966--1974, 2015.

\bibitem{kingma2014adam}
D.~P. Kingma and J.~Ba.
\newblock Adam: A method for stochastic optimization.
\newblock {\em arXiv preprint arXiv:1412.6980}, 2014.

\bibitem{kingma2013auto}
D.~P. Kingma and M.~Welling.
\newblock Auto-encoding variational bayes.
\newblock {\em arXiv preprint arXiv:1312.6114}, 2013.

\bibitem{kokic2017affordance}
M.~Kokic, J.~A. Stork, J.~A. Haustein, and D.~Kragic.
\newblock Affordance detection for task-specific grasping using deep learning.
\newblock In {\em 2017 IEEE-RAS 17th International Conference on Humanoid
  Robotics (Humanoids)}, pages 91--98. IEEE, 2017.

\bibitem{koppula2013learning}
H.~S. Koppula, R.~Gupta, and A.~Saxena.
\newblock Learning human activities and object affordances from rgb-d videos.
\newblock {\em The International Journal of Robotics Research}, 32(8):951--970,
  2013.

\bibitem{levine2016learning}
S.~Levine, P.~Pastor, A.~Krizhevsky, and D.~Quillen.
\newblock Learning hand-eye coordination for robotic grasping with large-scale
  data collection.
\newblock In {\em International Symposium on Experimental Robotics}, pages
  173--184. Springer, 2016.

\bibitem{lopes2007affordance}
M.~Lopes, F.~S. Melo, and L.~Montesano.
\newblock Affordance-based imitation learning in robots.
\newblock In {\em 2007 IEEE/RSJ International Conference on Intelligent Robots
  and Systems}, pages 1015--1021. IEEE, 2007.

\bibitem{lovchik1999robonaut}
C.~Lovchik and M.~A. Diftler.
\newblock The robonaut hand: A dexterous robot hand for space.
\newblock In {\em Proceedings 1999 IEEE international conference on robotics
  and automation (Cat. No. 99CH36288C)}, volume~2, pages 907--912. IEEE, 1999.

\bibitem{lowe2004distinctive}
D.~G. Lowe.
\newblock Distinctive image features from scale-invariant keypoints.
\newblock {\em International journal of computer vision}, 60(2):91--110, 2004.

\bibitem{maitin2010cloth}
J.~Maitin-Shepard, M.~Cusumano-Towner, J.~Lei, and P.~Abbeel.
\newblock Cloth grasp point detection based on multiple-view geometric cues
  with application to robotic towel folding.
\newblock In {\em 2010 IEEE International Conference on Robotics and
  Automation}, pages 2308--2315. IEEE, 2010.

\bibitem{manuelli2019kpam}
L.~Manuelli, W.~Gao, P.~Florence, and R.~Tedrake.
\newblock kpam: Keypoint affordances for category-level robotic manipulation.
\newblock {\em arXiv preprint arXiv:1903.06684}, 2019.

\bibitem{mayo20093d}
M.~Mayo and E.~Zhang.
\newblock 3d face recognition using multiview keypoint matching.
\newblock In {\em 2009 Sixth IEEE International Conference on Advanced Video
  and Signal Based Surveillance}, pages 290--295. IEEE, 2009.

\bibitem{mian2008keypoint}
A.~S. Mian, M.~Bennamoun, and R.~Owens.
\newblock Keypoint detection and local feature matching for textured 3d face
  recognition.
\newblock {\em International Journal of Computer Vision}, 79(1):1--12, 2008.

\bibitem{miller2011parametrized}
S.~Miller, M.~Fritz, T.~Darrell, and P.~Abbeel.
\newblock Parametrized shape models for clothing.
\newblock In {\em 2011 IEEE International Conference on Robotics and
  Automation}, pages 4861--4868. IEEE, 2011.

\bibitem{mousavian2019graspnet}
A.~Mousavian, C.~Eppner, and D.~Fox.
\newblock 6-dof graspnet: Variational grasp generation for object manipulation.
\newblock {\em arXiv preprint arXiv:1905.10520}, 2019.

\bibitem{nebehay2014consensus}
G.~Nebehay and R.~Pflugfelder.
\newblock Consensus-based matching and tracking of keypoints for object
  tracking.
\newblock In {\em IEEE Winter Conference on Applications of Computer Vision},
  pages 862--869. IEEE, 2014.

\bibitem{newell2016stacked}
A.~Newell, K.~Yang, and J.~Deng.
\newblock Stacked hourglass networks for human pose estimation.
\newblock In {\em European conference on computer vision}, pages 483--499.
  Springer, 2016.

\bibitem{nguyen2016detecting}
A.~Nguyen, D.~Kanoulas, D.~G. Caldwell, and N.~G. Tsagarakis.
\newblock Detecting object affordances with convolutional neural networks.
\newblock In {\em 2016 IEEE/RSJ International Conference on Intelligent Robots
  and Systems (IROS)}, pages 2765--2770. IEEE, 2016.

\bibitem{osiurak2010grasping}
F.~Osiurak, C.~Jarry, and D.~Le~Gall.
\newblock Grasping the affordances, understanding the reasoning: toward a
  dialectical theory of human tool use.
\newblock {\em Psychological review}, 117(2):517, 2010.

\bibitem{park2008multiple}
Y.~Park, V.~Lepetit, and W.~Woo.
\newblock Multiple 3d object tracking for augmented reality.
\newblock In {\em Proceedings of the 7th IEEE/ACM International Symposium on
  Mixed and Augmented Reality}, pages 117--120. IEEE Computer Society, 2008.

\bibitem{pinto2016supersizing}
L.~Pinto and A.~Gupta.
\newblock Supersizing self-supervision: Learning to grasp from 50k tries and
  700 robot hours.
\newblock In {\em 2016 IEEE international conference on robotics and automation
  (ICRA)}, pages 3406--3413. IEEE, 2016.

\bibitem{qi2017pointnet}
C.~R. Qi, H.~Su, K.~Mo, and L.~J. Guibas.
\newblock Pointnet: Deep learning on point sets for 3d classification and
  segmentation.
\newblock {\em Proc. Computer Vision and Pattern Recognition (CVPR), IEEE},
  2017.

\bibitem{ridge2010self}
B.~Ridge, D.~Sko{\v{c}}aj, and A.~Leonardis.
\newblock Self-supervised cross-modal online learning of basic object
  affordances for developmental robotic systems.
\newblock In {\em 2010 IEEE International Conference on Robotics and
  Automation}, pages 5047--5054. IEEE, 2010.

\bibitem{rublee2011orb}
E.~Rublee, V.~Rabaud, K.~Konolige, and G.~R. Bradski.
\newblock Orb: An efficient alternative to sift or surf.
\newblock In {\em ICCV}, volume~11, page~2. Citeseer, 2011.

\bibitem{csahin2007afford}
E.~{\c{S}}ahin, M.~{\c{C}}akmak, M.~R. Do{\u{g}}ar, E.~U{\u{g}}ur, and
  G.~{\"U}{\c{c}}oluk.
\newblock To afford or not to afford: A new formalization of affordances toward
  affordance-based robot control.
\newblock {\em Adaptive Behavior}, 15(4):447--472, 2007.

\bibitem{seita2018robot}
D.~Seita, N.~Jamali, M.~Laskey, R.~Berenstein, A.~K. Tanwani, P.~Baskaran,
  S.~Iba, J.~Canny, and K.~Goldberg.
\newblock Robot bed-making: Deep transfer learning using depth sensing of
  deformable fabric.
\newblock {\em arXiv preprint arXiv:1809.09810}, 2018.

\bibitem{tobin2017domain}
J.~Tobin, R.~Fong, A.~Ray, J.~Schneider, W.~Zaremba, and P.~Abbeel.
\newblock Domain randomization for transferring deep neural networks from
  simulation to the real world.
\newblock In {\em 2017 IEEE/RSJ International Conference on Intelligent Robots
  and Systems (IROS)}, pages 23--30. IEEE, 2017.

\bibitem{toshev2014deeppose}
A.~Toshev and C.~Szegedy.
\newblock Deeppose: Human pose estimation via deep neural networks.
\newblock In {\em Proceedings of the IEEE conference on computer vision and
  pattern recognition}, pages 1653--1660, 2014.

\bibitem{toussaint2018differentiable}
M.~Toussaint, K.~Allen, K.~A. Smith, and J.~B. Tenenbaum.
\newblock Differentiable physics and stable modes for tool-use and manipulation
  planning.
\newblock In {\em Robotics: Science and Systems}, 2018.

\bibitem{tulsiani2015viewpoints}
S.~Tulsiani and J.~Malik.
\newblock Viewpoints and keypoints.
\newblock In {\em Proceedings of the IEEE Conference on Computer Vision and
  Pattern Recognition}, pages 1510--1519, 2015.

\bibitem{van2010gravity}
J.~Van Den~Berg, S.~Miller, K.~Goldberg, and P.~Abbeel.
\newblock Gravity-based robotic cloth folding.
\newblock In {\em Algorithmic Foundations of Robotics IX}, pages 409--424.
  Springer, 2010.

\bibitem{washburn1960tools}
S.~L. Washburn.
\newblock Tools and human evolution.
\newblock {\em Scientific American}, 203(3):62--75, 1960.

\bibitem{wu2016single}
J.~Wu, T.~Xue, J.~J. Lim, Y.~Tian, J.~B. Tenenbaum, A.~Torralba, and W.~T.
  Freeman.
\newblock Single image 3d interpreter network.
\newblock In {\em European Conference on Computer Vision}, pages 365--382.
  Springer, 2016.

\bibitem{xie2019improvisation}
A.~Xie, F.~Ebert, S.~Levine, and C.~Finn.
\newblock Improvisation through physical understanding: Using novel objects as
  tools with visual foresight.
\newblock {\em arXiv preprint arXiv:1904.05538}, 2019.

\bibitem{zhu2015understanding}
Y.~Zhu, Y.~Zhao, and S.~Chun~Zhu.
\newblock Understanding tools: Task-oriented object modeling, learning and
  recognition.
\newblock In {\em Proceedings of the IEEE Conference on Computer Vision and
  Pattern Recognition}, pages 2855--2864, 2015.

\end{thebibliography}
}

\end{document}